\pdfoutput=1

\documentclass[11pt]{article}

\usepackage[]{acl}

\usepackage{times}
\usepackage{latexsym}
\usepackage{textcomp}
\usepackage{float}
\usepackage{multirow}
\usepackage{booktabs}
\usepackage{graphicx}
\usepackage[T1]{fontenc}
\usepackage{courier}
\usepackage{mathtools}
\usepackage{url}

\usepackage[utf8]{inputenc}
\usepackage{subfig}
\usepackage[belowskip=-5pt,aboveskip=5pt]{caption}

\usepackage{microtype}
\usepackage{inconsolata}
\usepackage[figuresright]{rotating}

\usepackage{tgcursor}
\usepackage{xcolor}
\usepackage{xspace}
\usepackage[utf8]{inputenc}
\usepackage{authblk}
\usepackage{natbib}
\usepackage{lipsum}
\usepackage{booktabs, 
            multirow, threeparttable}
\usepackage{enumitem}

\newtheorem{definition}{Definition}

\usepackage{rotating}
\providecommand{\lb}{\mbox{$\langle$}}
\providecommand{\rb}{\mbox{$\rangle$}}

\title{\textsc{GenRES}: Rethinking Evaluation for Generative Relation Extraction in the Era of Large Language Models}

\author[ ]{\textbf{Pengcheng Jiang}}
\author[ ]{\textbf{Jiacheng Lin}}
\author[ ]{\textbf{Zifeng Wang}}
\author[ ]{\textbf{Jimeng Sun}}
\author[ ]{\textbf{Jiawei Han}}
\affil[ ]{Department of Computer Science, University of Illinois at Urbana-Champaign}
\affil[ ]{\texttt{\{pj20,jl254,zifengw2,jimeng,hanj\}@illinois.edu}}

\begin{document}
\maketitle
\begin{abstract}

The field of relation extraction (RE) is experiencing a notable shift towards generative relation extraction (GRE), leveraging the capabilities of large language models (LLMs). However, we discovered that traditional relation extraction (RE) metrics like precision and recall fall short in evaluating GRE methods. This shortfall arises because these metrics rely on exact matching with human-annotated reference relations, while GRE methods often produce diverse and semantically accurate relations that differ from the references. To fill this gap, we introduce \textsc{GenRES} for a multi-dimensional assessment in terms of the topic similarity, uniqueness, granularity, factualness, and completeness of the GRE results. With \textsc{GenRES}, we empirically identified that (1) precision/recall fails to justify the performance of GRE methods; (2) human-annotated referential relations can be incomplete; (3) prompting LLMs with a fixed set of relations or entities can cause hallucinations. Next, we conducted a human evaluation of GRE methods that shows \textsc{GenRES} is consistent with human preferences for RE quality. Last, we made a comprehensive evaluation of fourteen leading LLMs using \textsc{GenRES} across document, bag, and sentence level RE datasets, respectively, to set the benchmark for future research in GRE\footnote{Source code and guidelines are available at \url{https://github.com/pat-jj/GenRES}}. 
\end{abstract}

\begin{figure}[!ht]
    \centering
    \includegraphics[width=1.0\linewidth]{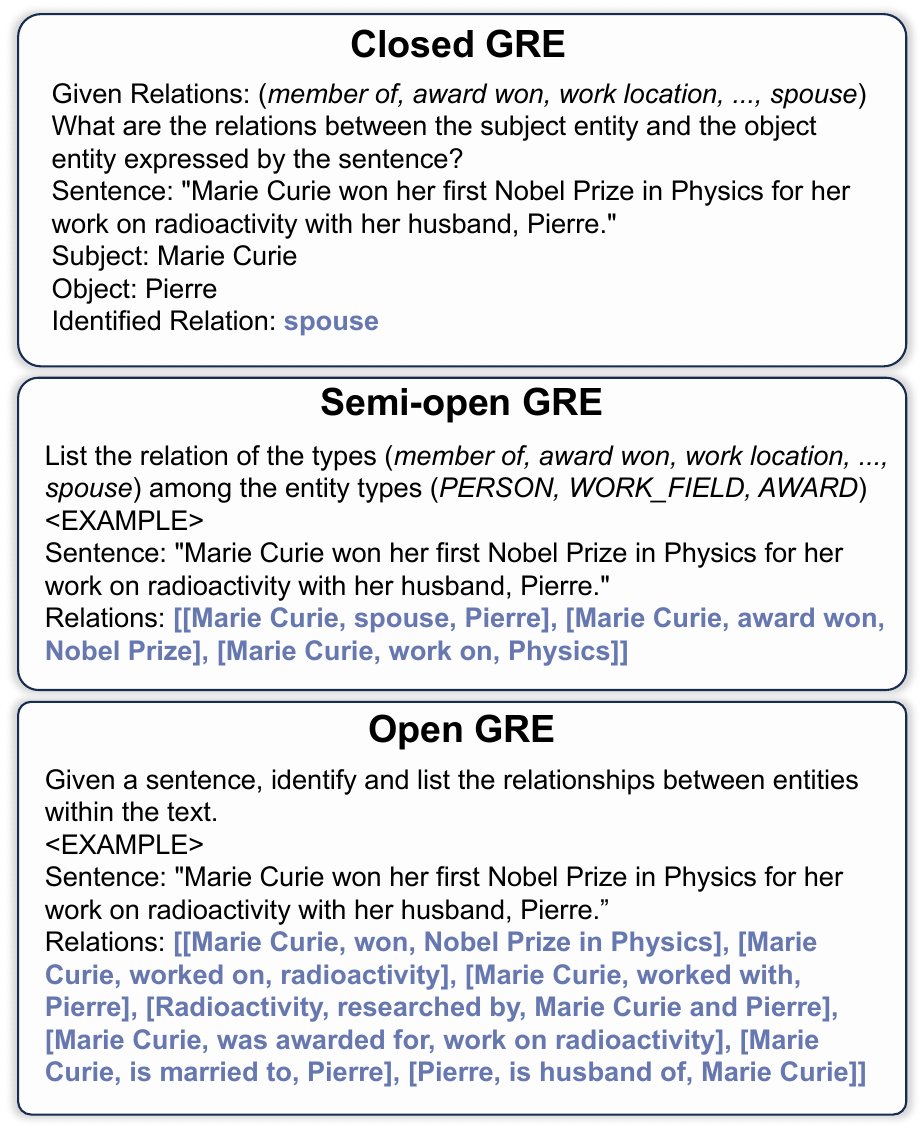}
    \caption{\textbf{Generative Relation Extraction (GRE):} Contrasting Closed and Semi-open GRE's type constraints with Open GRE's reliance on source text alone.}
    \label{fig:opengre}
    \vspace{-1em}
\end{figure}
\section{Introduction}
Relation Extraction (RE) is one of the most critical tasks in natural language processing \cite{han-etal-2020-data}. In essence, RE transforms unstructured text into structured, actionable knowledge (e.g., knowledge graphs). However, the traditional RE methods only mine the predefined patterns referring to the predefined sets of relations and entities, thus often struggling to capture the complexity of natural language. Recently, Large Language Models (LLMs) like GPT \cite{DBLP:journals/corr/abs-2303-08774}, promise a transition to Generative Relation Extraction (GRE).
LLM-based GRE methods are capable of comprehending the input texts and then identifying complex relationships without the constraints of predefined patterns in a zero-shot manner. This is particularly advantageous when there is a scarcity of training data, and the input texts are varied.

Existing applications of LLMs in GRE are either performing binary classification tasks \cite{li2023revisiting} given entity pairs and a set of predefined relation types, or given restricted entity types \cite{wadhwa-etal-2023-revisiting, zhu2023llms}, which overlook extensive novel relations and entities beneath the text. Notably, to unlock the full power of LLMs in GRE, we advocate a transformation from ``defining a set of relation types'' $\rightarrow$ ``finding matches between entities'' to ``exploring as many relations and entities as possible without limitation'' $\rightarrow$ ``refinement'' \cite{ paulheim2017knowledge, 10.1145/3186727}. This strategy elicits LLMs’ implicit knowledge to discover a wider array of relationships with minimal predefined constraints \cite{hao-etal-2023-bertnet}, which we define as ``Open GRE'' that can be applied to knowledge graph construction for various downstream tasks \cite{baralis2013graphsum, 10.1145/3178876.3186175, mohamed2020discovering, zeng2022toward, jiang2023graphcare}. We illustrate the difference of GRE strategies in Figure~\ref{fig:opengre}.

The versatility of GRE, however, poses significant challenges in evaluation \cite{wadhwa-etal-2023-revisiting}. Specifically, we identified that traditional relation extraction (RE) metrics like precision and recall only capture the exact matching with human-annotated reference relations, while GRE methods often produce diverse and semantically accurate relations that differ from the references. As such, we argue that precision in GRE should be verified against the source text, and recall should be based on soft matching to accommodate the output flexibility of generative models. Furthermore, a proficient model should not only cover crucial information in the text but also avoid redundant results, ensuring the extracted knowledge is both comprehensive and atomistic. To navigate these new dimensions, we introduce \textbf{\textsc{GenRES}} (\textbf{\textsc{Gen}}erative \textbf{\textsc{R}}elation \textbf{\textsc{E}}xtraction \textbf{\textsc{S}}coring), a multi-dimensional framework tailored for evaluating GRE. Our key contributions are as follows.
\vspace{-1em}
\begin{itemize}[leftmargin=*]
\setlength\itemsep{0.1em}
    \item We demonstrate the effectiveness of \textsc{GenRES} for evaluating GRE tasks, emphasizing its superiority over traditional metrics.
    \item We benchmark the open GRE performance of fourteen leading LLMs through \textsc{GenRES}, and paving the way for future research and development of better LLM-based GRE methods.
\end{itemize}



\section{Preliminaries}
\begin{definition} [\textbf{Source Document}]
    A source document $\mathcal{D}$ is a piece of free-text, which can be a sentence, a passage, or a document.
\end{definition}

\begin{definition} [\textbf{Extracted Triples}]
    A triple $\tau =$ $\lb s | r | o\rb$ is a structure formatting a piece of free text into a subject $s$, a relation $r$, and an object $o$. Example: For a sentence "Alice lives in Champaign.", "Alice" is the subject, "live in" is the relation, and "Champaign" is the object. Together, they form a triple $\lb Alice | live\_in | Champaign\rb$. We define $\mathcal{T}_{\mathcal{D}} = [\tau_1, \tau_2 ,...]$ as a list of triples extracted from the source document $\mathcal{D}$.
\end{definition}

\subsection{Generative Relation Extraction}
GRE uses a generative large language model (LLM) to extract relational triples from a source document $\mathcal{D}$. The model functions on an autoregressive basis at the token level, expressed as $P(x_{t} | x_{1}, x_{2}, \ldots, x_{t-1}, \mathcal{D})$, where $x_{t}$ represents the $t^{th}$ token in the output sequence. The process generates a sequence of tokens that are structured into triples $\mathcal{T}_{\mathcal{D}} = [\tau_{1}, \tau_{2}, \ldots]$. We categorize existing GRE methods as follows:
\begin{itemize}[leftmargin=*]
\setlength\itemsep{0.1em}
    \item \textbf{Closed GRE} \cite{li2023revisiting}: Given (1) source context,  (2) entity pairs in the context, and (3) a set of predefined relation types, prompt the LLM to classify the relation type between the entity pairs to compose each triple $\tau_i$.
    \item \textbf{Semi-open GRE} \cite{wadhwa-etal-2023-revisiting}: Given (1) source context, (2) a predefined set of relation types, and (3) a predefined set of entity types, prompt the LLM to extract triples $\tau_i$.
    \item \textbf{Open GRE}: Given source context, prompt the LLM to extract triples as many as possible.
\end{itemize}


\begin{figure*}[!h]
    \centering
    \includegraphics[width=\linewidth]{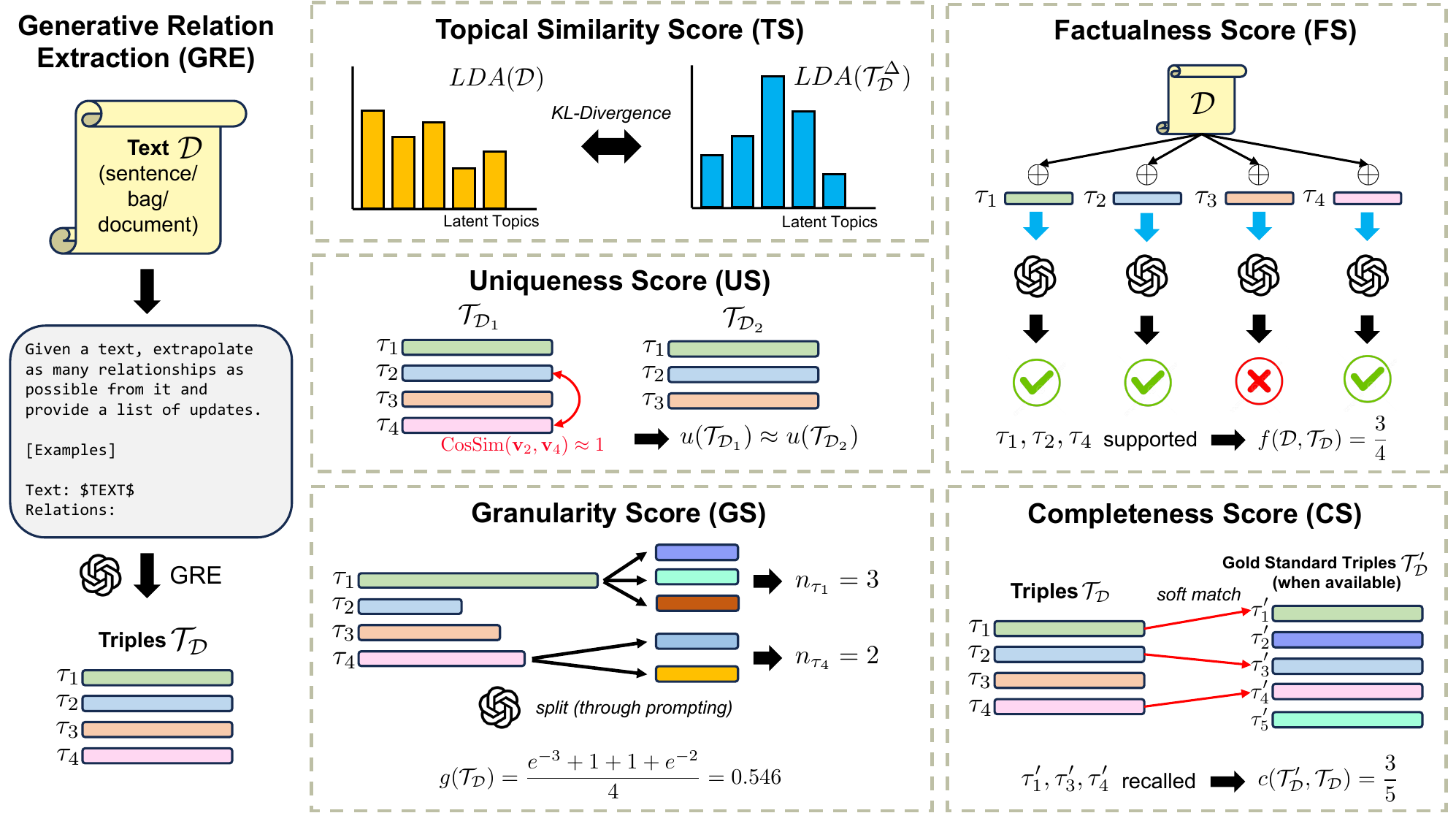}
    \caption{\textbf{\textsc{GenRES} framework for the evaluation of generative relation extraction (GRE).} \textbf{\textit{Left}}: An example showing the GRE process to extract triples $\mathcal{T}_{\mathcal{D}}$ from a source text $\mathcal{D}$ through prompting generative large language model. \textit{\textbf{Right}}: illustration of sub-scores contained in GREScore regarding: Topical Similarity (\S \ref{subsec:TS}), Uniqueness (\S \ref{subsec:US}), Fatualness (\S \ref{subsec:FS}), Granularity (\S \ref{subsec:GS}), and Completeness (\S \ref{subsec:CS}). }
    \label{fig:framework}
    \vspace{-0.5em}
\end{figure*}
\section{\textsc{GenRES}}
\label{sec:genres}
Evidenced by previous work conducting semi-open GRE \cite{wadhwa-etal-2023-revisiting}, traditional metrics for RE like hard matching precision/recall/F1 are inadequate to evaluate GRE tasks as the LLM generations are flexible. To fill in this gap, we introduce \textsc{GenRES}, an automated multi-aspect evaluation framework for GRE. \textsc{GenRES} are composed of a series of sub-scores defined as follows.

\subsection{Topical Similarity Score}
\label{subsec:TS}
We compute the topical similarity score (TS) to measure the information abundance of the extracted triples $\mathcal{T}_{\mathcal{D}}$ compared to the source text $\mathcal{D}$. Here, we employ a Latent Dirichlet Allocation (LDA) model \cite{blei2003latent}, an algorithm that represents each document as a blend of a certain number of latent topics, for topic modeling. We concatenate the elements in each triple so that $\mathcal{T}^{\Delta}_{\mathcal{D}} = [\tau_1', \tau_2', ...] = [s_1 \oplus r_1 \oplus o_1, s_2 \oplus r_2 \oplus o_2, ...]$. TS is computed as:
\begin{equation}
    t(\mathcal{D}, \mathcal{T}^{\Delta}_{\mathcal{D}}) = e^{-\sum_{i=1}^{K} LDA(\mathcal{D})_i
    \cdot \log \left( \frac{LDA(\mathcal{D})_i}{LDA(\mathcal{T}^{\Delta}_{\mathcal{D}})_i} \right)}
\end{equation}
which is based on the \textit{KL-divergence} of two topical distributions. A higher TS indicates that the extracted triples closely align with the topical content of the source document, reflecting effective and relevant information extraction, while a lower TS suggests that the extracted triples may be missing key topical elements from the source.

\subsection{Uniqueness Score}
\label{subsec:US}
Uniqueness Score (US) assesses the diversity of the extracted triples $\mathcal{T}_{\mathcal{D}}$ in the GRE, emphasizing the importance of extracting varied and distinct relationships. Given $\mathcal{T}_{\mathcal{D}} = [\tau_1, \tau_2, \ldots, \tau_n]$, with each triple $\tau_i$ encoded in a vector $\mathbf{v}_i$ using word embeddings, the US is computed as follows:
\begin{equation}
    u(\mathcal{T}_{\mathcal{D}}) = \frac{1}{n(n-1)} \sum_{i=1}^{n} \sum_{j \neq i}^{n} \left(\text{CosSim}(\mathbf{v}_i, \mathbf{v}_j) < \phi \right)
\end{equation}
where $\text{CosSim}(\mathbf{v}_i, \mathbf{v}_j)$ is the cosine similarity between the vector representations of triples $\tau_i$ and $\tau_j$. $\phi$ is a predefined similarity threshold. The normalization factor \( n(n-1) \) accounts for all pairings where \( i \neq j \). A higher US indicates greater diversity among the triples, while a lower US suggests more similarity and potential redundancy. 

\subsection{Factualness Score}
\label{subsec:FS}
Factualness Score (FS) quantifies the extent to which extracted triples, denoted as $\mathcal{T}_{\mathcal{D}}$, align with the information in the source text $\mathcal{D}$. This metric is crucial for gauging the hallucinations \cite{zhang2023sirens}, a phenomenon where LLMs fabricate the content not present in the source text. Building on the foundations laid by prior research \cite{min2023factscore, jiang-etal-2021-exploring-listwise}, FS employs a detailed triple-wise verification process. Each triple $\tau$ within $\mathcal{T}_{\mathcal{D}}$ undergoes a thorough check to confirm whether it is supported by factual evidence in $\mathcal{D}$:
\begin{equation}
f(\mathcal{D}, \mathcal{T}_{\mathcal{D}}) = \frac{1}{|\mathcal{T}_{\mathcal{D}}|} \sum_{\tau \in \mathcal{T}_{\mathcal{D}}} [\![\tau \text{ is supported by } \mathcal{D}]\!]
\end{equation}
where $[\![\tau \text{ is supported by } \mathcal{D}]\!]$  is an indicator function that returns $1$ if the triple is factual and $0$ if it is not. In this study, we adopt the approach from previous work \cite{min2023factscore} and utilize an LLM as the fact-checking tool. Specifically, we employ GPT-3.5-Turbo-Instruct as the fact checker, with the methodology detailed in Appendix \ref{ap:fact_prompt}. A high FS signifies that a substantial portion of the extracted triples are factually consistent with the source text. On the contrary, a low FS indicates a higher incidence of hallucinated or unsupported data. Employing this metric is vital to guarantee the reliability and trustworthiness of the information generated by the model.

\subsection{Granularity Score}
\label{subsec:GS}
The Granularity Score (GS) evaluates the level of detail of the extracted triples $\mathcal{T}_{\mathcal{D}}$ from the source text $\mathcal{D}$. It is based on the premise that triples should capture the optimal granularity of information, not too coarse. The GS aims to penalize triples that are overly broad and could be further split into more precise statements.
The process involves an assessment of each triple's potential to be split into more granular sub-triples. This can be performed by prompting an LLM to evaluate if a given triple can be divided into additional, more specific triples. The number of possible splits is represented by $n_{\tau}$ for each triple $\tau$.

The Granularity Score for the extracted triples $\mathcal{T}_{\mathcal{D}}$ is calculated using the formula:
\begin{equation}
    g(\mathcal{T}_{\mathcal{D}}) = \frac{1}{|\mathcal{T}_{\mathcal{D}}|} \sum_{\tau \in \mathcal{T}_{\mathcal{D}}} e^{-n_{\tau}}
\end{equation}
where \(e^{-n_{\tau}}\) is the exponential decay function based on the number of splits \(n_{\tau}\), which assigns a lower score to triples that can be split into more sub-triples (indicating they are too broad or general). Therefore, a lower Granularity Score indicates that the triples could be broken down further, while a higher score suggests that the triples are at an appropriate level of specificity. 

\subsection{Completeness Score}
\label{subsec:CS}
The Completeness Score (CS) evaluates how comprehensively the extracted triples $\mathcal{T}_{\mathcal{D}}$ cover the information present in the source text $\mathcal{D}$. This score is analogous to the recall metric in information retrieval and is particularly important when gold standard triples $\mathcal{T}'_{\mathcal{D}}$ are available for comparison.
CS is assessed by determining the proportion of gold standard triples that are successfully captured by the extracted triples. For each gold standard triple $\tau'$, we find the best matching triple $\tau$ from $\mathcal{T}_{\mathcal{D}}$, using cosine similarity of their embeddings as the \textit{soft matching} criterion. If the cosine similarity exceeds a specified threshold $\phi$, the triple $\tau$ is considered a match. CS is then computed as:
\begin{equation}
    c(\mathcal{T}'_{\mathcal{D}}, \mathcal{T}_{\mathcal{D}}) = \frac{|\{\tau' \in \mathcal{T}'_{\mathcal{D}} | \exists \tau \in \mathcal{T}_{\mathcal{D}}, \text{sim}(\tau, \tau') \geq \phi\}|}{|\mathcal{T}'_{\mathcal{D}}|}
\end{equation}
where $\text{sim}(\tau, \tau') = \text{CosSim}(emb(\tau), emb(\tau'))$ calculates the cosine similarity between the embeddings of the extracted triple and the gold standard triple. The threshold $\phi$ is pre-defined to determine the acceptable level of similarity for a match.
A higher CS indicates that the extracted triples effectively capture the complete range of information as represented by the ``gold standard''. It is worth noting that CS is optional as precise human annotations are expensive and not always available.

\section{Experiments}
\subsection{Datasets}
In our evaluation, we examine several RE datasets with a focus on their performance in GRE using test sets enriched with detailed human annotations. These include: \textbf{CDR} \cite{li2016biocreative}, a document-level dataset with 1,500 PubMed abstracts highlighting chemical-disease interactions; \textbf{DocRED} \cite{yao2019docred}, also document-level, derived from Wikipedia and Wikidata, featuring extensive entity, coreference, and relational annotations across 5,053 documents; \textbf{NYT10m} and \textbf{Wiki20m}  \cite{han-etal-2019-opennre}, both bag-level\footnote{A ``bag'' of information that share the same entity pair. 
} datasets from The New York Times and Wikipedia, respectively, with manually annotated test sets; and \textbf{TACRED} \cite{zhang2017position} and \textbf{Wiki80} \cite{han2018fewrel}, sentence-level datasets, the former comprising 106,264 examples across various text sources and the latter containing 56,000 instances with 80 relations from Wikipedia and Wikidata. These datasets collectively offer a comprehensive view of RE capabilities across various levels and sources.

We adopt a random sampling method to select the test sets from the above datasets. We randomly choose \{200, 500, 800\} samples for the document-, bag-, and sentence-level evaluations\footnote{For the Wiki20m dataset (bag-level), we deviated from this approach due to the predominance of low-quality random samples, often containing only a single ground-truth triple. We first refined the dataset to include samples with two triples, narrowing it down to 3,526 samples. From this filtered pool, 500 samples were randomly selected.}.

\begin{figure*}[t]
    \centering
    \includegraphics[width=\linewidth]{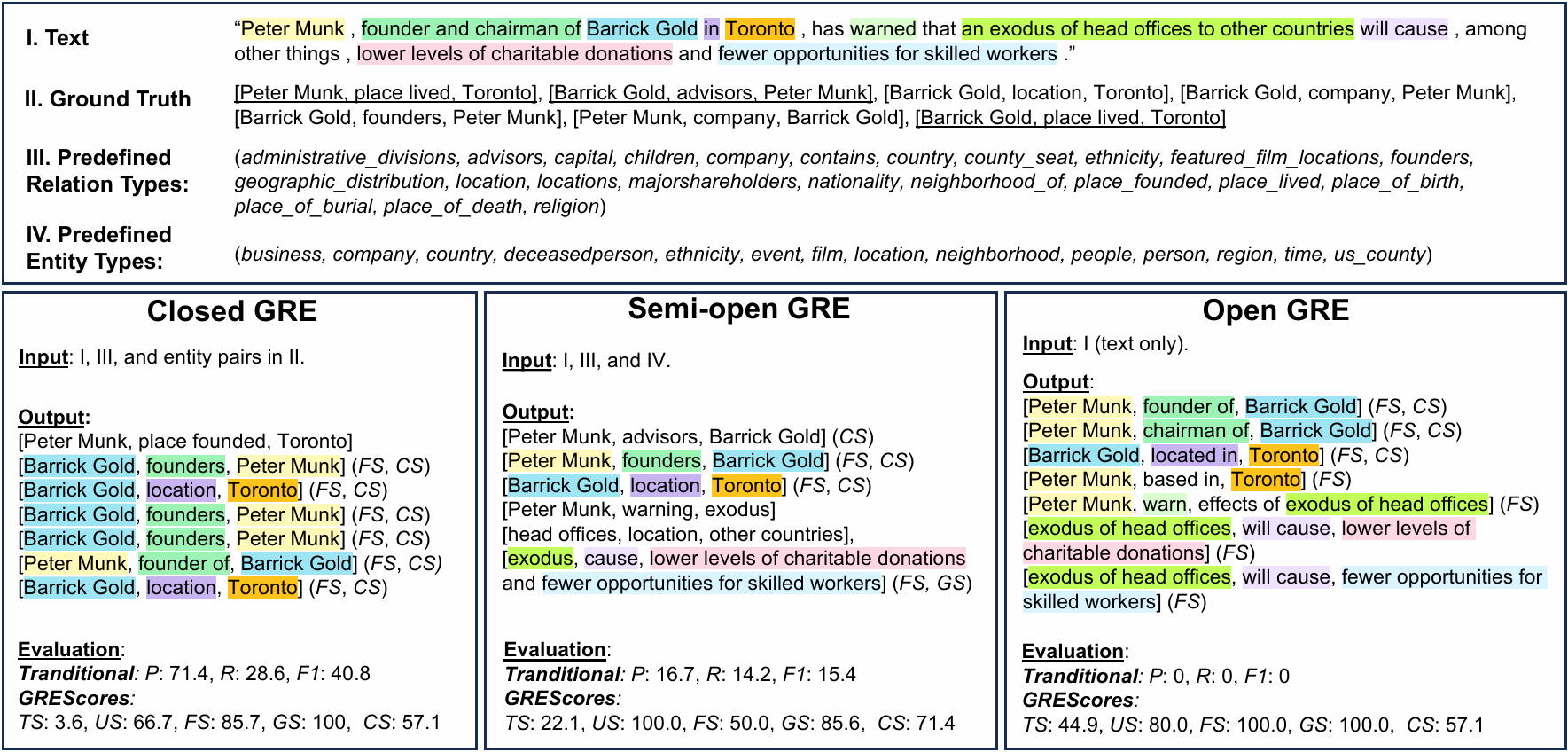}
    \caption{\textbf{Comparative Analysis of GRE Methods and Evaluation Metrics using the NYT10m Dataset.} The diagram showcases the outcomes of closed, semi-open, and open Generative Relation Extraction (GRE) strategies. The distinct entity and relation spans are color-coded, with factual triples specifically highlighted. The extracted triples that affect FS, CS (soft recall), and GS are listed with the corresponding labels. We \underline{underline} the ground truth labels that are inaccurate or cannot be inferred from the source text.}
    \label{fig:case_study}
    \vspace{-0.5em}
\end{figure*}
\subsection{Implementation}
For topical similarity score (TS), we train six LDA models with \{50, 100, 150, 150, 150, 150\} latent topic numbers and \{1500, 5051, 11086, 14257, 38140, 22400\} samples (document/bag/sentence) for CDR, DocRED, NYT10m, Wiki20m, TACRED, and Wiki80, respectively.
For evaluations (US and CS) using word embedding, we retrieve the embedding for each entity and relation in the triple using \texttt{text-embedding-ada-002}, and perform element-wise addition to obtain the triple embedding.\footnote{Concatenation should be employed instead when the direction of the relation is concerned}
Based on our tests, we set the similarity threshold $\phi$ at 0.95. All local LLMs are run on 8 NVIDIA A100 GPUs. All prompts used are detailed in Appendix \ref{ap:prompts}.

\begin{table}[t]
\small
\centering
\setlength{\tabcolsep}{3.9pt}
\resizebox{\linewidth}{!}{
\begin{threeparttable}
\begin{tabular}{lcccccccc}
\toprule
& \multicolumn{4}{c}{\textbf{CDR}}     & \multicolumn{4}{c}{\textbf{NYT10m}} \\
\cmidrule(lr){2-5} \cmidrule(lr){6-9} 
& \textbf{C}   &\textbf{S} & \textbf{O} &\textbf{GT} & \textbf{C}    & \textbf{S} & \textbf{O} &\textbf{GT} \\
\midrule
\textbf{\textit{\#tri}} & 10.1  & 6.8   & 16.1 & 10.1 &1.4    & 2.9   & 5.8   &1.4\\
\textbf{\textit{\#tok}} & 6.6   & 4.0   &8.3   & 5.8 &4.6    & 2.0   & 7.0    &4.5\\
\midrule
\textbf{\textit{P}}   &58.8 &1.1  &0.4    &-        &29.3       &5.2    &0.0    &- \\
\textbf{\textit{R}}   &58.7 &0.8  &0.7    &-         &26.6       &12.7  &0.0    &- \\
\textbf{\textit{F1}}  &58.8 &0.7  &0.5    &-        &27.5       &6.5    &0.0    &- \\
\midrule
\textbf{\textit{TS}} & 11.9 & 35.5  &\textbf{77.6} & 9.6 & 10.3   & 13.4  &\textbf{54.2}   &8.7\\
\textbf{\textit{US}} & 31.8 & 58.2  &\textbf{89.6} & 33.4  & 87.5  & \textbf{91.5}  &83.0  &69.3\\
\textbf{\textit{FS}} & 64.4 & 62.0  &\textbf{96.8} & 93.5 & 72.3  & 33.7  &\textbf{84.0}   &84.1\\
\textbf{\textit{GS}} & \textbf{92.0} & 78.5  &54.2 & 98.1 & \textbf{87.4}  & 79.9  &71.9   &93.1\\
\textbf{\textit{CS}} & \textbf{58.4}$^{*}$ & 56.7  &47.8 & 100 & \textbf{62.3}$^{*}$  & 20.3  &53.4    &100\\
\bottomrule

\end{tabular}
\begin{tablenotes}[flushleft]\footnotesize
\item $^*$Closed GRE, due to its use of predefined entity pairs for relation classification, inherently exhibits high triple similarity. Hence, we further check relation embedding similarity for the best soft matching of triples.
\end{tablenotes}

\caption{\textbf{Different GRE strategies measured by different metrics including traditional P/R/F1 and \textsc{GenRES}.} ``C'', ``S'', ``O'', and ``GT'' denote Closed, Semi-open, Open GRE, and ground truth, respectively. GPT-3.5-Turbo-Instruct was used as the LLM.  We \textbf{highlight} the highest GREScores for each dataset.}
\label{tb:diff_gres}
\end{threeparttable}}
\vspace{-1em}
\end{table}

\subsection{Performance of Different GRE Strategies}
We conducted evaluations of closed, semi-open, and open GRE on the CDR and NYT10m datasets. The expansive relation sets and the absence of defined entity types in other datasets render them incompatible with closed and semi-open GRE, owing to the limitations of context window constraints. This limitation emphasizes the flexibility of open GRE, which operates unconstrained by predefined relation types or entity types, proving its adaptability to a wider array of datasets. The comparative results of these evaluations are presented in Table \ref{tb:diff_gres}. Combined with our example shown in Figure \ref{fig:case_study}, we summarize the key observations as follows.

\noindent\textbf{Traditional metrics are not ideal for GRE evaluation}, especially in semi-open and open GRE settings. Figure 3 illustrates that despite open GRE's high-quality extractions based on FS and CS, they score zero across these metrics. This occurs because Precision/Recall/F1 depend on exact matching of triples, which are nearly impossible without predefined relation/entity sets, as evidenced by the zero scores for these metrics on the NYT10m dataset in Table \ref{tb:diff_gres}. This finding syncs with \citet{wadhwa-etal-2023-revisiting}'s conclusion.

\noindent\textbf{Human annotations sometimes are unreliable.} In Figure 3, we underline several mistakes (e.g., ``[Barrick Gold, advisors, Peter Munk], [Barrick Gold, place lived, Toronto]'') in the the ground truth where ``Barrick Gold'' is a company but incorrectly recognized as a person. Such inaccurate labels are unlikely to be correctly predicted by LLMs. This suggests that traditional metrics that purely rely on ground truth triples, are even inadequate for closed GRE, and more so for semi-open and open GRE.

\begin{table*}[t]
\small
\centering
\setlength{\tabcolsep}{3.9pt}
\resizebox{\textwidth}{!}{

\begin{tabular}{clcccccccccccccc}
\toprule
&& \multicolumn{7}{c}{\textbf{CDR}}     & \multicolumn{7}{c}{\textbf{DocRED}} \\
\cmidrule(lr){3-9} \cmidrule(lr){10-16} 

& &\textbf{\textit{\#tri}}  &\textbf{\textit{\#tok}}  & \textbf{\textit{TS}} & \textbf{\textit{US}} & \textbf{\textit{FS}} & \textbf{\textit{GS}}   & \textbf{\textit{CS}}
&\textbf{\textit{\#tri}}  &\textbf{\textit{\#tok}}  & \textbf{\textit{TS}} & \textbf{\textit{US}} & \textbf{\textit{FS}} & \textbf{\textit{GS}}   & \textbf{\textit{CS}}
\\ 
\midrule
& Ground Truth
& 10.1  & 5.8 & 9.6  & 33.4 & 93.5 & 98.1 & 100
& 12.4 & 6.0 & 8.4  & 64.0 & 94.4 & 81.9 & 100
\\
\midrule
\multirow{5}{*}{\rotatebox{0}{\textbf{LLaMA}}}
&Vicuna-7B 
& 6.8 & 8.4 & 57.8 & 86.9 & 84.7 & 44.6 &  30.7 
& 7.4 & 9.9 & 23.1 & 81.9 & 93.4 & 46.8 &  28.3 \\

&Vicuna-33B
& 6.4 & 10.5 & 73.0 & 89.2 & \textbf{97.3} & 38.4 &  32.0 
& 10.8 & 9.8 & 34.7 & 82.8 & 97.2 & 49.6 &  36.9 \\

&LLaMA-2-7B 
& 5.6 & 6.7 & 48.6 & 92.0 & 62.0 & 44.9 &  25.7 
& 2.7 & 3.2 & 12.8 & 93.3 & 34.0 & 60.6 &  12.1 \\

&LLaMA-2-70B  
& 10.8 & 8.1 & \textbf{74.8} & 87.6 & 96.6 & \textbf{57.8} &  \textbf{51.0} 
& 13.8 & 8.7 & \textbf{39.2} & 82.6 & \textbf{97.3} & \textbf{60.9} &  \textbf{39.2} \\

&WizardLM-70B 
& 10.2 & 7.8 & 65.4 & \textbf{94.1} & 76.4 & 46.2 &  32.6 
& 5.8 & 3.6 & 24.3 & \textbf{94.9} & 37.9 & 56.7 &  12.8 \\

\midrule
\multirow{5}{*}{\rotatebox{0}{\textbf{GPT}}}
&text-davinci-003 
& 12.7 & 8.3 & 76.7 & 87.2 & 96.8 & \textbf{55.4} &  44.3 
& 15.3 & 8.5 & 40.1 & 84.2 & 97.6 & 59.8 &  46.2 \\

&GPT-3.5-Turbo-Inst.
& 16.1 & 8.3 & 77.6 & 89.6 & 96.8 & 54.2 &  47.8 
& 17.8 & 8.9 & 47.8 & 85.6 & 98.1 & 56.2 &  44.7 \\

&GPT-3.5-Turbo 
& 11.2 & 11.4 & 81.7 & 89.2 & \textbf{98.2} & 40.3 &  30.2 
& 15.0 & 9.9 & \textbf{50.4} & 84.0 & 98.5 & 49.1 &  36.5 \\

&GPT-4 
& 14.3 & 9.3 & 81.7 & 91.0 & 97.9 & 49.1 &  46.3 
& 17.8 & 8.7 & 48.6 & 82.8 & \textbf{98.6} & 59.6 &  47.3 \\

&GPT-4-Turbo
& 18.6 & 8.5 & \textbf{82.1} & \textbf{91.9} & 96.8 & 53.1 &  \textbf{48.8} 
& 21.5 & 8.7 & 50.0 & \textbf{87.4} & 97.6 & \textbf{63.1} &  \textbf{49.3} \\

\midrule
\multirow{4}{*}{\rotatebox{0}{\textbf{others}}}
&Mistral-7B-Inst.
& 14.2 & 9.1 & 69.0 & 74.9 & 93.5 & 51.1 &  \textbf{40.0} 
& 11.3 & 9.6 & 30.2 & 76.4 & 94.1 & 55.2 &  27.5 \\

&Zephyr-7B-Beta 
& 25.9 & 8.8 & 49.1 & 79.5 & 70.1 & \textbf{57.7} &  29.3 
& 18.6 & 8.6 & 27.9 & 79.4 & 94.7 & \textbf{64.7} &  37.1 \\

&Galactica-30B
& 0.2 & 0.3 & 4.1 & 1.1 & 0.9 & 44.4 & 0.0 
& 0.0 & 0.0 & 8.6 & 0.0 & 0.0 & 0.0 & 0.0 \\

&OpenChat-3.5 
& 8.6 & 12.6 & \textbf{78.7} & \textbf{91.9} & \textbf{97.4} & 38.2 &  31.8 
& 15.4 & 8.9 & \textbf{39.7} & \textbf{82.1} & \textbf{98.1} & 61.7 &  \textbf{43.4} \\
\bottomrule
\end{tabular}}

\caption{\textbf{\textsc{GenRES} evaluation of Open GRE on \textit{document-level} datasets.} Scores (\%) are averaged across documents. \textit{\#tri} and \textit{\#tok} denote the number of triples per document and the number of tokens per triple, respectively. We \textbf{highlight} the highest within-group scores. Galactica's low scores are due to its limited size of context window.}
\label{tb:grescore_perf_doc}
\end{table*}

\begin{table*}[!h]
\small
\centering
\setlength{\tabcolsep}{3.9pt}
\resizebox{\textwidth}{!}{

\begin{tabular}{clcccccccccccccc}
\toprule
&& \multicolumn{7}{c}{\textbf{NYT10m}}     & \multicolumn{7}{c}{\textbf{Wiki20m}} \\
\cmidrule(lr){3-9} \cmidrule(lr){10-16} 

&&\textbf{\textit{\#tri}}  &\textbf{\textit{\#tok}}  & \textbf{\textit{TS}} & \textbf{\textit{US}} & \textbf{\textit{FS}} & \textbf{\textit{GS}}   & \textbf{\textit{CS}}
&\textbf{\textit{\#tri}}  &\textbf{\textit{\#tok}}  & \textbf{\textit{TS}} & \textbf{\textit{US}} & \textbf{\textit{FS}} & \textbf{\textit{GS}}   & \textbf{\textit{CS}}

\\ 
\midrule
&Ground truth 
& 1.4 & 4.5 & 8.7  & 69.3 & 84.1 & 93.1 & 100
& 2.0 & 6.3 & 4.4 & 21.2 & 88.7 & 85.1 & 100
\\
\midrule
\multirow{5}{*}{\rotatebox{0}{\textbf{LLaMA}}}
&Vicuna-7B 
& 3.1 & 7.8 & 42.0 & 86.4 & 80.0 & 60.2 & 38.9 
& 3.0 & 7.5 & 48.3 & 67.8 & 50.0 & 68.6 & 37.3 \\
&Vicuna-33B 
& 4.7 & 7.2 & \textbf{47.8} & 80.1 & 75.1 & 65.2 & 46.5 
& 4.1 & 7.0 & \textbf{49.8} & 56.4 & 84.4 & 75.4 & 46.1 \\
&LLaMA-2-7B 
& 3.1 & 6.0 & 35.4 & 82.2 & 78.9 & 69.2 & 38.4
& 3.1 & 6.3 & 37.9 & \textbf{73.8} & 73.4 & 75.6 & 36.0 \\
&LLaMA-2-70B 
& 5.0 & 6.9 & 45.4 & 83.0 & \textbf{81.7} & \textbf{71.8} & \textbf{52.4} 
& 4.1 & 6.9 & 45.2 & 62.0 & \textbf{87.1} & \textbf{78.4} & \textbf{50.2} \\
&WizardLM-70B 
& 4.4 & 4.2 & 30.5 & \textbf{88.9} & 43.9 & 68.9 & 27.6 
& 3.6 & 5.6 & 43.1 & 67.8 & 67.3 & 75.0 & 40.9 \\
\midrule
\multirow{5}{*}{\rotatebox{0}{\textbf{GPT}}}
&text-davinci-003 
& 4.9 & 7.1 & 50.6 & 81.4 & 85.8 & 69.3 & 52.6
& 3.7 & 8.2 & 51.8 & 56.9 & 91.3 & 73.3 & 43.5 \\
&GPT-3.5-Turbo-Inst. 
& 5.8 & 7.0 & 54.2 & 83.0 & 84.0 & \textbf{71.9} & 53.4 
& 4.8 & 7.7 & 54.0 & 60.3 & 90.1 & 78.9 & 43.8 \\
&GPT-3.5-Turbo 
& 4.1 & 6.2 & 43.3 & 82.3 & 68.2 & 62.8 & 29.8
& 3.6 & 7.7 & 48.2 & 61.8 & 80.2 & 72.7 & 32.5 \\
&GPT-4 
& 5.1 & 7.4 & 56.2 & 81.8 & 89.0 & 68.2 & 52.6
& 3.8 & 8.1 & \textbf{59.0} & 56.2 & \textbf{93.2} & 77.2 & 40.0 \\
&GPT-4-Turbo 
& 5.3 & 7.8 & \textbf{58.1} & \textbf{84.2} & \textbf{89.6} & 69.1 & \textbf{53.7}
& 4.2 & 7.6 & 56.4 & \textbf{62.0} & 92.4 & \textbf{81.2} & \textbf{52.7} \\
\midrule
\multirow{4}{*}{\rotatebox{0}{\textbf{others}}}
&Mistral-7B-Inst. 
& 5.7 & 7.4 & 40.6 & 77.6 & 75.4 & 62.9 & 36.5
& 4.0 & 6.9 & 43.3 & 57.0 & 83.6 & 69.9 & 40.1 \\
&Zephyr-7B-Beta 
& 7.8 & 7.2 & 36.5 & 80.8 & 64.9 & \textbf{73.8} & 47.0 
& 5.2 & 6.8 & 40.3 & \textbf{65.5} & 75.5 & \textbf{79.0} & 45.9 \\
&Galactica-30B 
& 8.3 & 8.7 & 29.7 & 48.4 & 52.4 & 60.6 & 37.0 
& 6.0 & 8.4 & 35.3 & 49.4 & 65.2 & 66.8 & 38.6 \\
&OpenChat-3.5 
& 5.2 & 7.2 & \textbf{54.0} & \textbf{84.7} & \textbf{84.3} & 69.7 & \textbf{55.3} 
& 4.3 & 7.0 & \textbf{57.5} & 61.8 & \textbf{90.5} & 76.0 & \textbf{47.7} \\
\bottomrule
\end{tabular}}

\caption{\textbf{\textsc{GenRES} evaluation of Open GRE on \textit{bag-level} datasets.} Scores (\%) are averaged across bags. \textit{\#tri} and \textit{\#tok} denote the number of triples per bag and the number of tokens per triple, respectively. We \textbf{highlight} the highest within-group scores.}
\vspace{-1em}

\label{tb:grescore_perf_bag}
\end{table*}

\begin{table*}[t]
\small
\centering
\setlength{\tabcolsep}{3.9pt}
\resizebox{\textwidth}{!}{

\begin{tabular}{clcccccccccccccc}
\toprule
&& \multicolumn{7}{c}{\textbf{TACRED}}     & \multicolumn{7}{c}{\textbf{Wiki80}} \\
\cmidrule(lr){3-9} \cmidrule(lr){10-16} 

&&\textbf{\textit{\#tri}}  &\textbf{\textit{\#tok}}  & \textbf{\textit{TS}} & \textbf{\textit{US}} & \textbf{\textit{FS}} & \textbf{\textit{GS}}   & \textbf{\textit{CS}}
&\textbf{\textit{\#tri}}  &\textbf{\textit{\#tok}}  & \textbf{\textit{TS}} & \textbf{\textit{US}} & \textbf{\textit{FS}} & \textbf{\textit{GS}}   & \textbf{\textit{CS}}
\\ 
\midrule
&Ground Truth
& 1.4 & 4.6 & 15.8 & 92.7 & 87.0 & 94.9 & 100
& 1.0 & 5.8 & 5.9 & 100 & 90.1 & 84.4 & 100
\\
\midrule
\multirow{5}{*}{\rotatebox{0}{\textbf{LLaMA}}}
&Vicuna-7B 
& 2.6 & 8.7 & 40.4 & 85.0 & \textbf{75.6} & 58.9 & 36.2 
& 2.4 & 7.9 & 41.3 & 76.8 & 81.0 & 61.7 & 36.6 \\
&Vicuna-33B 
& 4.3 & 7.3 & \textbf{44.3} & 75.5 & 71.0 & 69.2 & 47.2 
& 3.8 & 7.2 & \textbf{47.3} & 62.1 & 79.9 & 73.8 & 46.8 \\
&LLaMA-2-7B 
& 2.8 & 6.3 & 36.7 & 85.3 & 66.9 & 71.2 & 37.8 
& 2.4 & 5.8 & 25.8 & 69.8 & 60.4 & \textbf{76.9} & 31.4 \\
&LLaMA-2-70B 
& 4.1 & 6.4 & 40.8 & 79.3 & 74.5 & \textbf{76.8} & \textbf{56.4} 
& 3.7 & 6.6 & 41.5 & 64.8 & \textbf{82.4} & \textbf{76.9} & \textbf{49.4} \\
&WizardLM-70B 
& 2.1 & 2.9 & 23.3 & \textbf{90.7} & 28.0 & 72.1 & 9.8 
& 2.1 & 3.2 & 25.6 & \textbf{84.9} & 36.6 & 74.4 & 21.4 \\
\midrule
\multirow{5}{*}{\rotatebox{0}{\textbf{GPT}}}
&text-davinci-003 
& 4.4 & 7.1 & 56.1 & 79.8 & 84.0 & 72.8 & 58.6 
& 4.0 & 6.8 & 59.2 & 65.3 & 89.2 & 74.9 & 51.9 \\
&GPT-3.5-Turbo-Inst. 
& 5.0 & 7.0 & 58.6 & 80.5 & 81.6 & 72.6 & 58.6 
& 4.4 & 6.9 & 60.2 & 69.3 & 88.7 & \textbf{75.4} & \textbf{54.8} \\
&GPT-3.5-Turbo 
& 3.9 & 6.8 & 52.7 & 81.1 & 76.4 & 67.5 & 39.7 
& 3.4 & 6.3 & 50.9 & \textbf{69.5} & 75.6 & 68.9  & 36.0 \\
&GPT-4 
& 4.3 & 7.5 & \textbf{59.1} & 80.4 & 87.6 & 69.1 & 57.8 
& 4.0 & 7.1 & \textbf{65.4} & 66.2 & 92.3 & 74.2 & 47.8 \\
&GPT-4-Turbo 
& 4.4 & 7.8 & 58.5 & \textbf{82.6} & \textbf{88.6} & \textbf{73.2} & \textbf{63.4} 
& 4.0 & 7.6 & 61.9 & 69.4 & \textbf{92.8} & 74.5 & 47.1 \\
\midrule
\multirow{4}{*}{\rotatebox{0}{\textbf{others}}}
&Mistral-7B-Inst. 
& 4.7 & 7.1 & 43.9 & 78.6 & 71.0 & 65.5 & 41.2 
& 3.6 & 7.8 & 44.6 & 67.8 & 83.9 & 67.7 & 38.5 \\
&Zephyr-7B-Beta 
& 5.4 & 7.6 & 36.4 & 78.6 & 65.8 & 72.0 & 44.9 
& 4.5 & 7.8 & 43.2 & 68.1 & 77.8 & 74.2 & 42.6 \\
&Galactica-30B 
& 8.5 & 8.9 & 33.4 & 43.9 & 57.5 & 64.1 & 30.9 
& 5.6 & 7.2 & 35.0 & 47.9 & 63.1 & 73.3 & 38.4 \\
&OpenChat-3.5 
& 4.3 & 7.1 & \textbf{50.7} & \textbf{80.8} & \textbf{80.4} & \textbf{72.1} & \textbf{60.0} 
& 4.0 & 7.0 & \textbf{53.8} & \textbf{69.7} & \textbf{88.7} & \textbf{74.9} & \textbf{50.6} \\
\bottomrule
\end{tabular}}

\caption{\textbf{\textsc{GenRES} evaluation of Open GRE on \textit{sentence-level} datasets.} Scores (\%) are averaged across sentences. \textit{\#tri} and \textit{\#tok} denote the number of triples per sentence and the number of tokens per triple, respectively. We \textbf{highlight} the highest within-group scores.}

\label{tb:grescore_perf_sent}
\end{table*}

\begin{figure*}[h]
    \centering
    \includegraphics[width=1.0\linewidth]{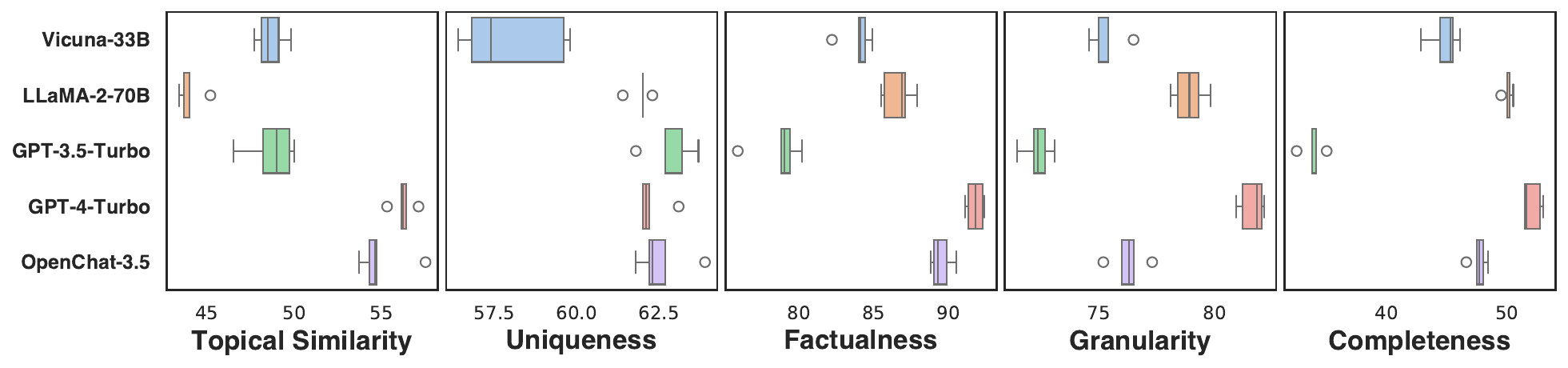}
    \caption{\textbf{GRE performance of five LLMs on Wiki20m, each with five runs with random seeds.}}
    \label{fig:gre_perf_multi_runs}
    \vspace{-0.5em}
\end{figure*}
\noindent\textbf{The imposition of predefined relation sets or entity types can misguide LLMs to generate inaccurate triples.} For instance, as seen in Figure \ref{fig:case_study}, closed GRE misclassifies the relation between ``Peter Munk'' and ``Toronto'' as ``place founded'' based on limited choices from the relation set, despite the text not supporting this inference. Similarly, semi-open GRE's entity recognition becomes problematic when it erroneously divides ``exodus of head offices'' into separate entities ``exodus'' and ``head offices'', leading to less coherent and less meaningful triples. 

It is also obvious that the range of information captured by extracted triples widens from closed GRE to open GRE. Closed and semi-open GRE, which limit the types of relations or entities, often yield extractions with a narrower scope. This constriction hampers the completeness of the captured information, a fact corroborated by the TS metrics presented in Table \ref{tb:diff_gres}. Furthermore, providing a more diverse relation set to semi-open GRE, such as the one in NYT10m (as opposed to the more limited CDR, which restricts entity types to chemicals and diseases), results in a significant drop in granularity (GS). In contrast, open GRE maintains stability, underscoring the benefit of eschewing predefined relation/entity types. Although closed GRE records the highest GS and CS, it is benefited from taking extra input entity pairs, which are not provided to simi-open and open GRE.

\subsection{Open GRE Performance of LLMs}
Due to the aforementioned advantages of Open GRE, we further test the capabilities of the leading LLMs to perform this task, which includes \textbf{LLaMA Family} \cite{touvron2023llama1, touvron2023llama2}: LLaMA-2-7B, LLaMA-2-70B, Vicuna-1.5-7B, Vicuna-1.3-33B, and WizardLM-70B \cite{xu2023wizardlm}. \textbf{GPT Family} \cite{brown2020language}: text-davinci-003, GPT-3.5-Turbo (1106), GPT-3.5-Turbo-Instruct, GPT-4, and GPT-4-Turbo \cite{DBLP:journals/corr/abs-2303-08774}. \textbf{Others}: Mistral-7B-Instruct \cite{DBLP:journals/corr/abs-2310-06825}, Zephyr-7B-Beta \cite{tunstall2023zephyr}, GALACTICA \cite{taylor2022galactica}, and OpenChat-3.5 \cite{wang2023openchat}. Models are selected majorly based on their performance on Chatbot Arena \cite{zheng2023judging}. Our evaluation results are shown in Tables \ref{tb:grescore_perf_doc}, \ref{tb:grescore_perf_bag}, and \ref{tb:grescore_perf_sent}. 

\begin{figure*}[!t]
    \centering
    \includegraphics[width=1.0\linewidth]{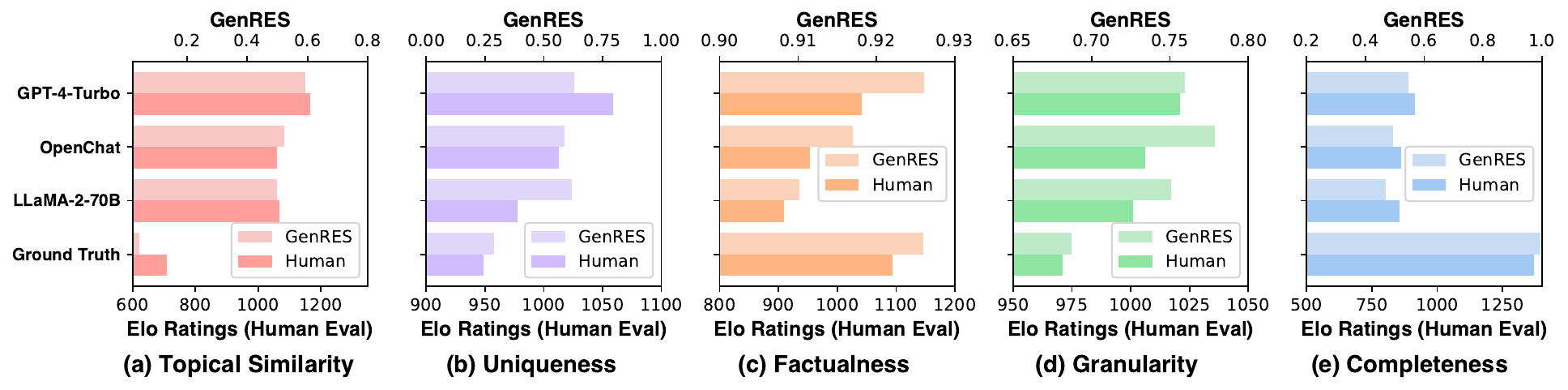}
    \caption{\textbf{Human Preference Evaluation (Elo Ratings) vs GenRES Evaluation on 100 Wiki20m samples.}}
    \label{fig:human_eval}
    \vspace{-0.5em}
\end{figure*}
We summarize our findings as follows.

(1) Within individual datasets, LLaMA-2-70B, GPT-4-Turbo, and OpenChat emerge as the top performers in their respective categories based on the highest scores obtained across six datasets. Inter-dataset comparisons reveal that the GPT family consistently outperforms others in Topical Similarity (TS), likely due to their supreme capability to interpret the full content of the text unit. Surprisingly, a light model - OpenChat-3.5 (7B) ourperforms heavier LLMs like Galactica-30B, Vicuna-33B, LLaMA-2-70B, WizardLM-70B, text-davinci-003, and GPT-3.5-Turbo on most datasets.

(2) High Completeness Score (CS) can indicate high Factualness Score (FS). This means human annotations are still valuable to evaluate GRE with our soft matching recall. However, high FS does not indicate high CS, as Open GRE is not limited to the fixed relation/entity types.  We also observe that the factualness of GPT-4 and GPT-4-Turbo are consistently higher than that of ground truth.

(3) A greater number of tokens per triple does not inherently result in a lower Granularity Score (GS). This suggests that the GS metric can encourage models to identify more atomic relationships rather than merely focusing on brevity.

(4) We observed no clear correlation between the number of triples, Topical Similarity (TS), and Uniqueness Similarity (US), indicating the distinct significance of each metric. For instance, on the CDR dataset, Mistral-7B-Instruct and Zephyr-7B-Beta show that a larger output of triples does not necessarily equate to higher TS or lower US. While Zephyr-7B-Beta produces more off-topic triples than Mistral-7B-Instruct, it does not result in more repetitive content. This highlights the importance of evaluating each metric independently.

Figure \ref{fig:gre_perf_multi_runs} shows the GRE task performance of five leading LLMs tested with five random seeds on the Wiki20m dataset. The results demonstrate the models' high-quality generation and the effectiveness of our multi-dimensional evaluation framework. Notably, the models' consistent performance across different runs validates our nuanced evaluation metrics, highlighting their robustness in assessing GRE model performance.

Figure \ref{fig:human_eval} showcases the Elo Rating \cite{elo1978rating} results of 100 samples from Wiki20m dataset via human annotation and our proposed \textsc{GenRES}. In most cases, the model ranks by \textsc{GenRES} are consistent with human annotators. We also evaluate the consistency between human annotators using the tie-discounted accuracy \cite{gao2023examining}. We find the following agreement scores: Topical Similarity 81.0\%, Uniqueness 93.0\%, Factualness 82.7\%, Granularity 92.7 \%, and Completeness 88.2\%. These results showcase the consistency between the human annotators. More details of human evaluation can be found in Appendix \ref{sec:app_human_eval}.



\section{Related Works}
\paragraph{Open RE.} Open RE uncovers new relation types in unsupervised open-domain corpora, traditionally through tagging-based and clustering-based approaches. Tagging-based Open RE treats the task as sequence labeling, extracting relational phrases from sentences \cite{DBLP:journals/corr/abs-1908-01761, DBLP:conf/acl/CuiWZ18, DBLP:conf/naacl/StanovskyMZD18}, while clustering-based methods utilize external linguistic tools to feature-rich relations and cluster them into distinct types \cite{10.1162/tacl_a_00604, DBLP:journals/tacl/MarcheggianiT16, DBLP:conf/esws/ElSaharDGGL17}. With the rapid development of LLMs, recent work has demonstrated the effectiveness of LLMs in Open RE from a generative perspective \cite{wadhwa2023revisiting, li2023revisiting}. Our proposed \textsc{GenRES} focuses on Generative RE, bridging the existing gap in evaluating Open Generative RE techniques.

\paragraph{Generative RE.} Generative models have exhibited significant promise in the field of RE \cite{wadhwa2023revisiting, wan2023gpt, li2023revisiting}.  Sequence-to-sequence models such as BART \cite{DBLP:conf/acl/LewisLGGMLSZ20} were utilized to extract triples from input texts \cite{DBLP:journals/corr/abs-2202-13229, DBLP:conf/iclr/PaoliniAKMAASXS21, DBLP:conf/emnlp/CabotN21}. Then, LLMs were proved to be able to make zero-shot and few-shot generative RE without fine-tuning \cite{wadhwa2023revisiting,li2023revisiting}. Specifically, \citet{wadhwa2023revisiting} compared GPT-3 \cite{brown2020language} and FLAN-T5 \cite{DBLP:journals/corr/abs-2210-11416} to fully supervised RE methods and identified LLMs reach comparable performance in the zero-shot setup. However, existing GRE methods still rely on a predefined set of relations and entities similar to traditional RE. In this paper, we explore a more open setting and propose a unified evaluation framework \textsc{GenRES} applicable to all types of generative RE.

\paragraph{Evaluation for Text Generation.} The evaluation of text generation quality is central to benchmarking the performance of LLMs. While traditional metrics like BLEU \cite{DBLP:conf/acl/PapineniRWZ02} and ROUGE \cite{lin2004rouge} assess surface-level word matching, they often inadequately capture the quality of the generated text. BERTScore \cite{zhang2019bertscore} focuses on semantic similarity, but still missing the multifaceted nature of text generation. Recently, LLMs have been utilized to evaluate text generation quality, such as FActScore \cite{min2023factscore} on verifying the factualness, and UniEval \cite{zhong2022towards} on multi-aspect evaluation. In addition, GPTScore \cite{fu2023gptscore} utilizes LLMs for token-level probability analysis, enhancing flexibility in text assessment. Recent studies \cite{liu2023gpteval,gao2023human, li2023prd} explore prompting-based multi-aspect evaluation, broadening the scope of evaluation methods. Unlike all the above works, our \textsc{GenRES} is the first metric designed specifically for Generative RE tasks.

\section{Conclusions}
In this paper, we introduced \textsc{GenRES}, a framework for evaluating Generative Relation Extraction using Large Language Models, marking a significant shift in the NLP field. Our findings based on extensive tests highlight the potential of LLMs to transform relation extraction and set the stage for future research, potentially revolutionizing information extraction processes and applications across various domains.

\clearpage
\bibliography{references}

\clearpage
\appendix
\section{Limitations, Ethics, and Risks}
\subsection{Limitations}
\noindent \textbf{LLMs as Evaluators.} Within \textsc{GenRES}, we employ the GPT-3.5-Turbo-Instruct large language model (LLM) for assessing the factualness and granularity of extracted relationship triples. However, challenges arise when the LLM delivers incorrect evaluations, particularly in instances where information is overly implicit, misleading, debatable \cite{chen-etal-2019-seeing}, or when the model encounters its inherent hallucination issues \cite{zhang2023sirens}. To mitigate these problems, potential solutions include instructing the model to detail its reasoning process leading to a prediction \cite{wei2022chain}, or applying ensemble methods \cite{li2023revisiting} to determine the most likely answer. These approaches are areas of interest for our future research endeavors.

\noindent \textbf{Unfocused Extraction by Open GRE.} Our research champions the Open Generative Relation Extraction (Open GRE) paradigm, which motivates LLMs to harvest a broader array of relationships, unconstrained by specific relation or entity types. While this approach has demonstrated enhanced topical breadth and factual content in extractions, it also results in a less focused extraction process compared to traditional methods like closed GRE and semi-open GRE \cite{wadhwa2023revisiting, li2023revisiting}. For instance, in constructing a Knowledge Graph (KG) for medical question answering, certain extractions, such as the triple \lb John, age, 16\rb, might be irrelevant and hence undesirable for inclusion in the KG. However, we posit that an intermediary layer, such as post-processing, should exist between Relation Extraction (RE) and downstream applications. This step would serve to refine and tailor the extracted relationships to meet specific requirements, aligning with methodologies proposed in existing literature \cite{paulheim2017knowledge, 10.1145/3186727}. Moreover, our \textsc{GenRES} framework is versatile enough to assess all forms of GRE, with the Open GRE configuration, noted for its flexibility, serving as a particularly effective benchmark for evaluating the robustness of our approach.

\subsection{Ethics and Risks}
All datasets used in this study, namely CDR \cite{li2016biocreative}, DocRED \cite{yao2019docred}, NYT10m \cite{han-etal-2019-opennre}, Wiki20m \cite{han-etal-2019-opennre}, TACRED \cite{zhang2017position}, and Wiki80 \cite{han2018fewrel} are publicly available. This transparency minimizes ethical concerns related to data sourcing and usage.  

Additionally, the interpretability and transparency of LLM decision-making processes are paramount, particularly in contexts involving sensitive or personal data. Recognizing the limitations and error tendencies of LLMs, including occasional information inaccuracies, we emphasize the importance of reliability in our evaluation methods. Furthermore, the integration of LLMs as evaluators impacts traditional human roles, calling for a careful examination of the ethical implications of labor displacement. Lastly, the potent capabilities of LLMs underscore the need for responsible use and measures to prevent misuse, aligning our research with high ethical standards and societal well-being. We carefully checked and ensured that there is no offensive information contained in the data we used as the input to any LLMs.

\section{Templates for Prompting LLMs}
\label{ap:prompts}
\subsection{Templates for Generative Relation Extraction}
This appendix delineates the structured prompts and demonstrations utilized in our generative relation extraction methodology. The templates are devised to prime the model for precise and contextually relevant relationship extraction from textual data across different domains and levels of granularity.

\begin{figure*}[t]
    \centering
    \includegraphics[width=1.0\linewidth]{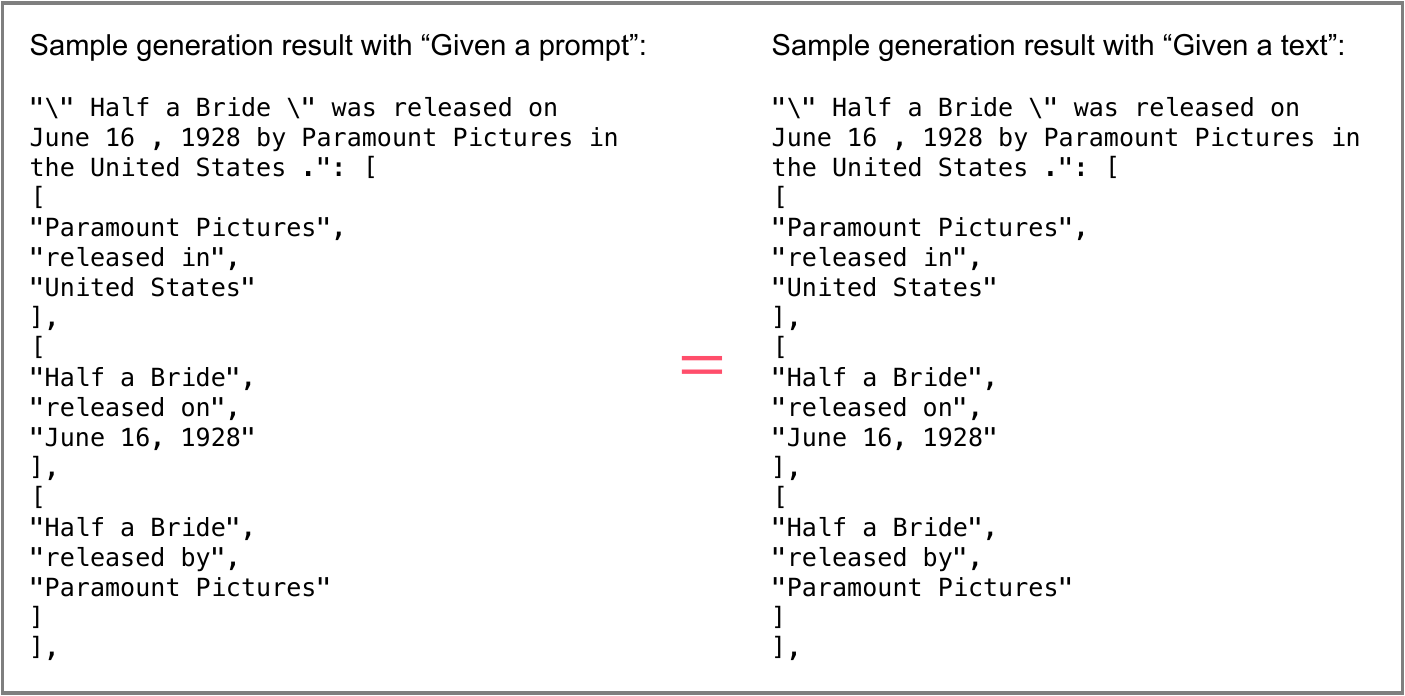}
    \caption{Comparison of Generation Using Different Templates: One Employing "Given a prompt" and the Other "Given a text" (Model: LLaMA-2-70B). The outcomes are identical.}
    \label{fig:diff_prompt_text}
\end{figure*}
\noindent \textbf{General Instruction\footnote{The term "Given a prompt" was initially utilized here instead of "Given a text", following the convention in GraphGPT: \url{https://github.com/varunshenoy/GraphGPT/blob/main/public/prompts/stateless.prompt}. The adjustment was also applied in a new experimental setting, where it produced triple generation results consistent with the original template, thanks to the in-context learning we applied. Hence, the term "text" is deemed more appropriate for our context. We showcase an example of identical generation in Figure \ref{fig:diff_prompt_text}.}:} The model is instructed to identify relationships between entities, with the aim to extract both intra-sentence and inter-sentence relational triples. This ensures a comprehensive understanding of the text, reflecting the intricacies of document-level nuances and the succinctness of sentence-level information.

\noindent \textbf{LLaMA-2 Model Instruction:} An additional directive is provided to the LLaMA-2 model to maintain output stability. The goal is to have the model generate a consistent list of triples, avoiding any extraneous information that does not contribute to the relationship representation.

\noindent \textbf{Demonstration Examples:} Examples are tailored to the general and biomedical domains to pre-heat the model towards the target topics. This stratagem is intended to:

\begin{itemize}[leftmargin=*]
    \item Facilitate the model's adaptation to the domain-specific language and context, thus enabling more accurate and relevant extractions.
    \item Encourage the model to discern and replicate the desired output structure from the examples, which is crucial for reliable relationship extraction.
\end{itemize}

The provided demonstrations span a variety of contexts and exemplify the format in which the relationships should be presented. The clear and topic-oriented examples aim to fine-tune the model's performance, ensuring it can navigate the complexities of relation extraction with precision across both biomedical and general domains.

\subsection{Template for Factualness Verification}
\label{ap:fact_prompt}
In the context of evaluating the factual accuracy of information extracted by language models, we present our template for factualness verification in Figure \ref{fig:fact_prompt}. Utilizing GPT-3.5-Turbo-Instruct as the language model evaluator, our template is designed to solicit a binary output: ``true'' if the relationship (triplet) is factually correct, ``false'' otherwise, based solely on the information entailed in the source text.

The template is constructed with three examples, each serving a specific purpose to calibrate the model's understanding of factual correspondence:

\begin{itemize}[leftmargin=*]
    \item \textbf{Example 1} establishes the model's ability to recognize direct factual statements that are explicitly stated in the source text.
    \item \textbf{Example 2} tests the model's discernment of geographical facts and common knowledge, challenging it to detect misinformation.
    \item \textbf{Example 3} assesses the model's capacity to correctly interpret narrative contexts and character relationships, a more subtle and complex form of factual verification.
\end{itemize}

The inclusion of these examples in the template aims to ensure that the model is thoroughly vetted across a spectrum of factual verification scenarios ranging from straightforward fact-checking to the interpretation of literary works.

\subsection{Template for Granularity Checking}
\label{ap:gran_prompt}
For granularity checking, we employ the template shown in Figure \ref{fig:gran_prompt}. The template contains 9 examples, to teach the LLM (GPT-3.5-Turbo-Instruct) what triples can be further split and what are not. Explanations are required when a triple cannot be split (GS = 0).

\begin{table*}[!h]
\small
\centering
\resizebox{0.7\linewidth}{!}{
\begin{tabular}{ll}
\toprule
\multicolumn{1}{c}{\textbf{Hyperparameter}} & \multicolumn{1}{c}{\textbf{Values}} \\
\midrule
\textbf{LDA latent topics} \\
CDR & \{20, 30, 40, \textbf{50}, 60, 70, 80, 90, 100\} \\ 
DocRED & \{30, 50, 70, \textbf{100}, 150\} \\ 
NYT10m & \{50, 100, \textbf{150}, 200, 250\} \\
Wiki20m & \{50, 100, \textbf{150}, 200, 250\}\\
TACRED & \{100, \textbf{150}, 200, 250, 300\}\\
Wiki80 & \{100, \textbf{150}, 200, 250, 300\}\\
\midrule
Triple similarity threshold $\phi$      & \{0.85, 0.90, 0.91, 0.92, 0.93, 0.94, \textbf{0.95}, 0.96, 0.97, 0.98\} \\
\midrule
\textbf{Open-source LLMs-related} \\
max\_new\_tokens & min[\#token\_limit, \{3, 5, 6, 7, \textbf{8}, 9, 10\}*\#input\_tokens] \\
floating-point number & \textbf{float16} \\
\midrule
\textbf{GPT-related}\\
max\_new\_tokens  & \textbf{800} \\
temperature & \textbf{0.3} \\
\bottomrule
\end{tabular}
}
\caption{\textbf{Hyperparameters Tuning.} We highlight the optimal ones based on our experiments in \textbf{bold}.}

\label{tb:hyper}
\end{table*}
\begin{figure}[!ht]
    \centering
    \vspace{2em}
    \includegraphics[width=1.0\linewidth]{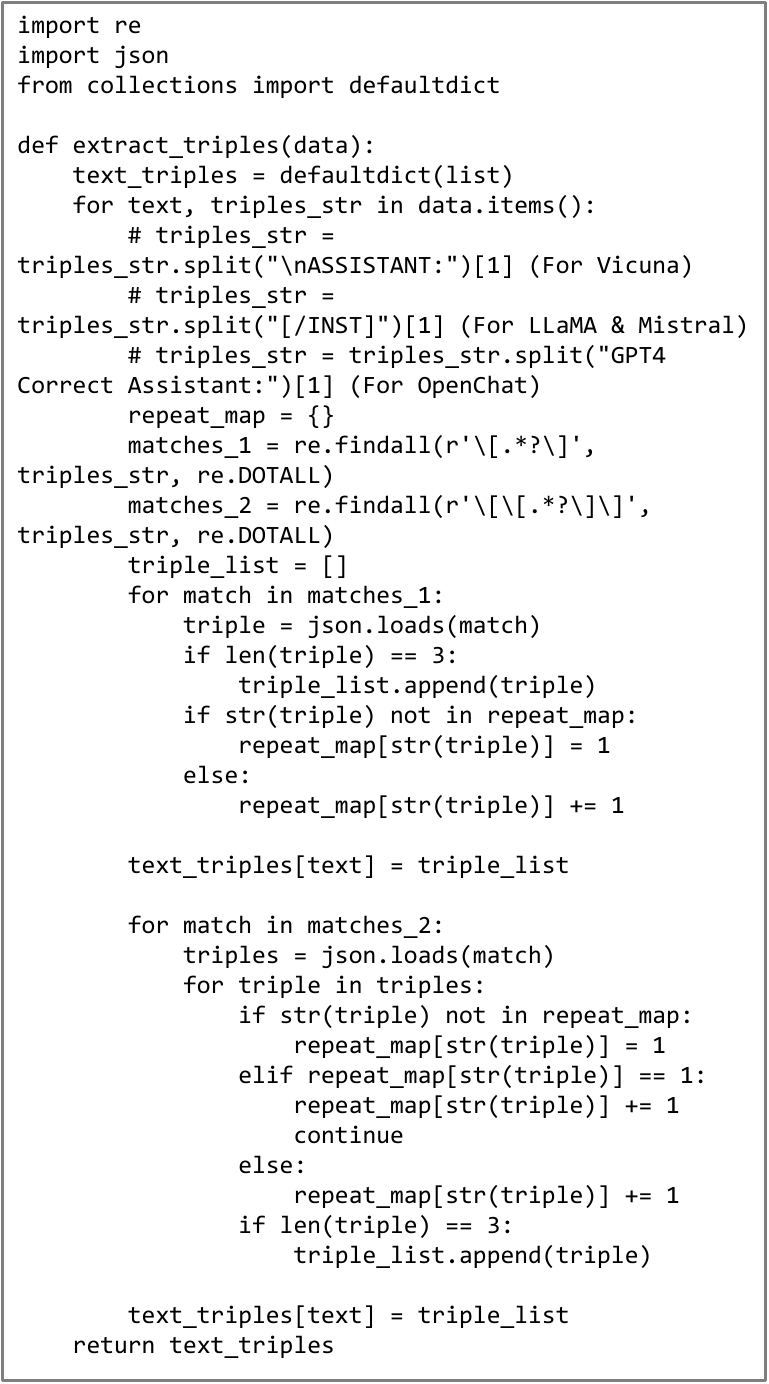}
    \caption{Post-processing of LLMs' generation.}
    \label{fig:post_pro}
\end{figure}
\section{Post-processing}
Figure \ref{fig:post_pro} displays the `extract\_triples` function, a Python-based method used for post-processing in our study. This function parses strings containing encoded triples in various formats (Vicuna, LLaMA \& Mistral, OpenChat) and extracts them as structured data. Using regular expressions, it identifies triples enclosed in single or double square brackets, decodes them via JSON. The result is a dictionary where each text is associated with a list of its extracted triples.

\section{Hyper-parameter Tuning}
The process of hyper-parameter tuning is crucial for optimizing the performance of our models. Table \ref{tb:hyper} presents a comprehensive list of the hyper-parameters adjusted during our experiments. This includes the number of latent topics for LDA, various dataset-specific parameters, and thresholds for triple similarity. Furthermore, specific parameters related to open-source LLMs and GPT-related configurations are tuned to enhance model efficiency and output quality.

\begin{figure}[!h]
    \centering
    \includegraphics[width=1.0\linewidth]{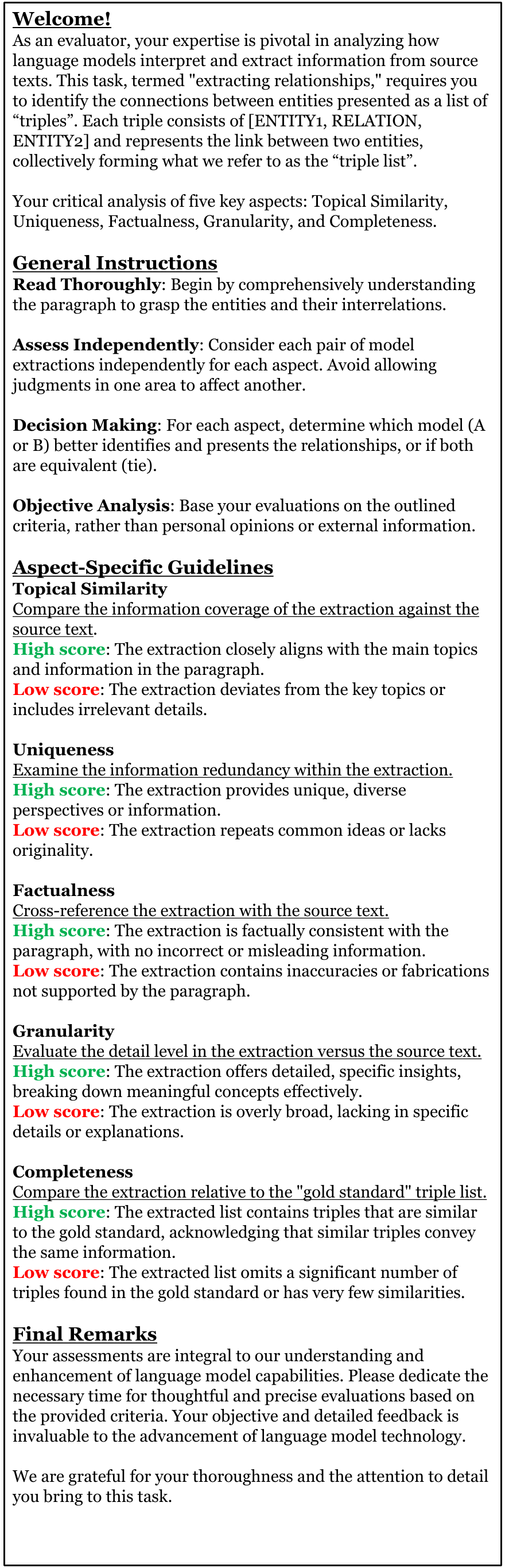}
    \caption{\textbf{Instruction for Human Annotators.}}
    \label{fig:human_inst}
\end{figure}
\section{Human Evaluation}
\label{sec:app_human_eval}
We further conducted human evaluation experiments to verify the alignment of our proposed GenRES with human preferences. Three annotators, who are all computer science graduate students, are involved in this evaluation.
\subsection{Evaluation Setup}
Our setup for human evaluation follows the approach detailed in studies such as \citet{gao2023examining}, \citet{DBLP:journals/corr/abs-2305-11206}, and \citet{dettmers2023qlora}. We adopt a pairwise comparison method for assessing model outputs. This approach simplifies the evaluation process by requiring human annotators to choose the better result from a pair of options. The evaluation was performed using 100 samples from the Wiki20m dataset. In this process, for each score proposed in Section \ref{sec:genres}, three human annotators compared the output relationships from Groundtruth, LLaMA-2-70b, OpenChat, and GPT-4-Turbo in pairs, leading to three possible outcomes for each pair: model A being superior, model B being superior, or a tie.  Subsequently, we apply the Elo rating \cite{elo1978rating} system to score the final results.

\paragraph{Elo Rating} Elo rating, initially established as a prevalent system for assessing player skill in chess and various competitive games, has recently been adapted to evaluate LLMs\footnote{\url{https://lmsys.org/blog/2023-05-03-arena/}} \cite{gao2023examining, DBLP:journals/corr/abs-2305-11206, dettmers2023qlora}. Its adaptability, characterized by features such as scalability and incremental adjustment, makes it particularly suitable for this purpose. This innovative use of the Elo rating system offers a robust quantitative framework for comparing the performance of various LLMs. In our pairwise comparison setup, the outcome of each comparison impacts the models' scores: a tie results in no change in scores, while a victory leads to an increase in the winner's score and a decrease in the loser's score. Following the completion of all comparisons, the Elo Rating system outputs a final score for each model, thereby establishing their relative rankings based on performance.

\paragraph{Instructions for Annotators} The instructions for annotators are shown in Figure \ref{fig:human_inst}. Annotators should evaluate the outputs from five aspects in Section \ref{sec:genres}. During the evaluation process, the models are anonymous for annotators. It should be noted that Completeness is measured after all other metrics have been assessed to prevent the leakage of ground truth information to annotators.

\paragraph{Inter-Annotator Agreement} To evaluate Inter-Annotator Agreement with tie-discounted accuracy, we randomly select 50 samples from the 100 Wiki20m samples, resulting in a total of 1500 overlap pairs for two human annotators.  This process aimed to assess the consistency level between annotators, anticipating a significant alignment in their evaluations. For the final scoring, we merged all the annotations. The scoring protocol for merging is as follows: (1) When both annotators' responses were in agreement, this consensus was accepted as the merged result. (2) If one annotator declared a tie, the decision of the other was taken as the final annotation. (3) If one annotator believed that ’model A wins’ and the other that ’model B wins,’ the models were considered tied.
\begin{figure*}[!h]
    \centering
    \includegraphics[width=1.0\linewidth]{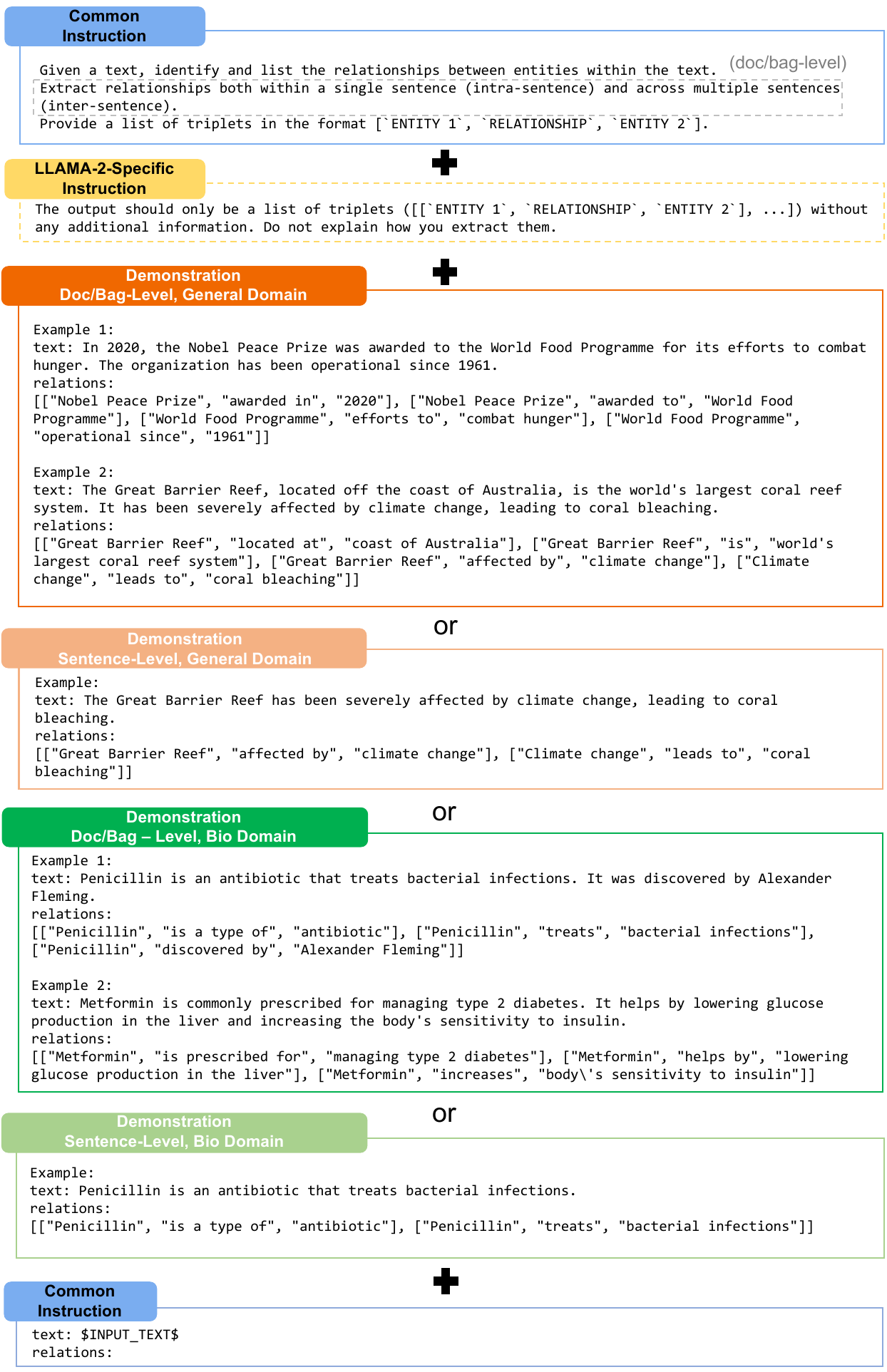}
    \caption{\textbf{Templates used for Open Generative Relation Extraction.}}
    \label{fig:gre_prompts}
\end{figure*}
\begin{figure*}[h]
    \centering
    \includegraphics[width=1.0\linewidth]{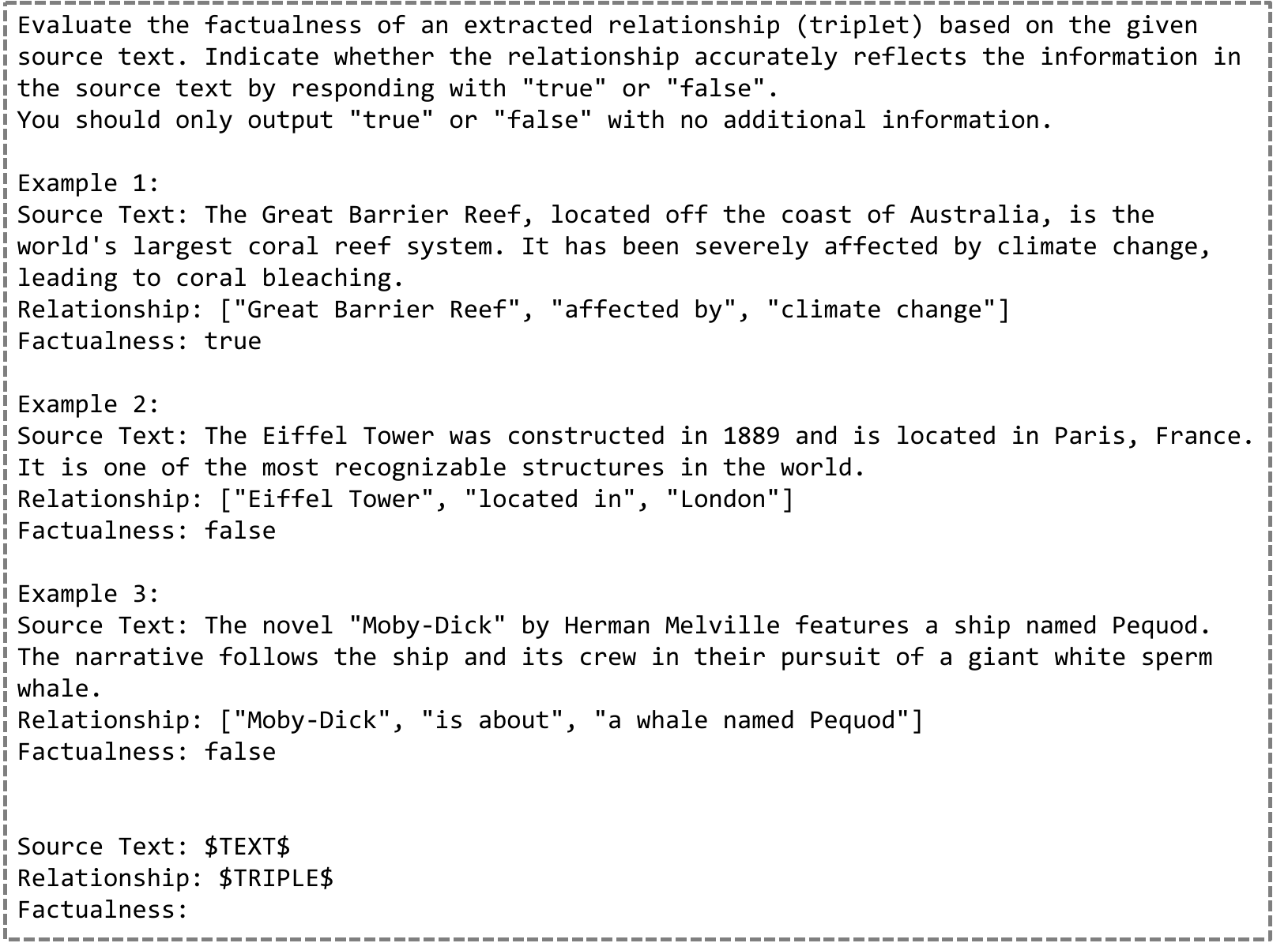}
    \caption{\textbf{Template for Factualness Verification.}}
    \label{fig:fact_prompt}
\end{figure*}
\begin{figure*}[h]
    \centering
    \includegraphics[width=1.0\linewidth]{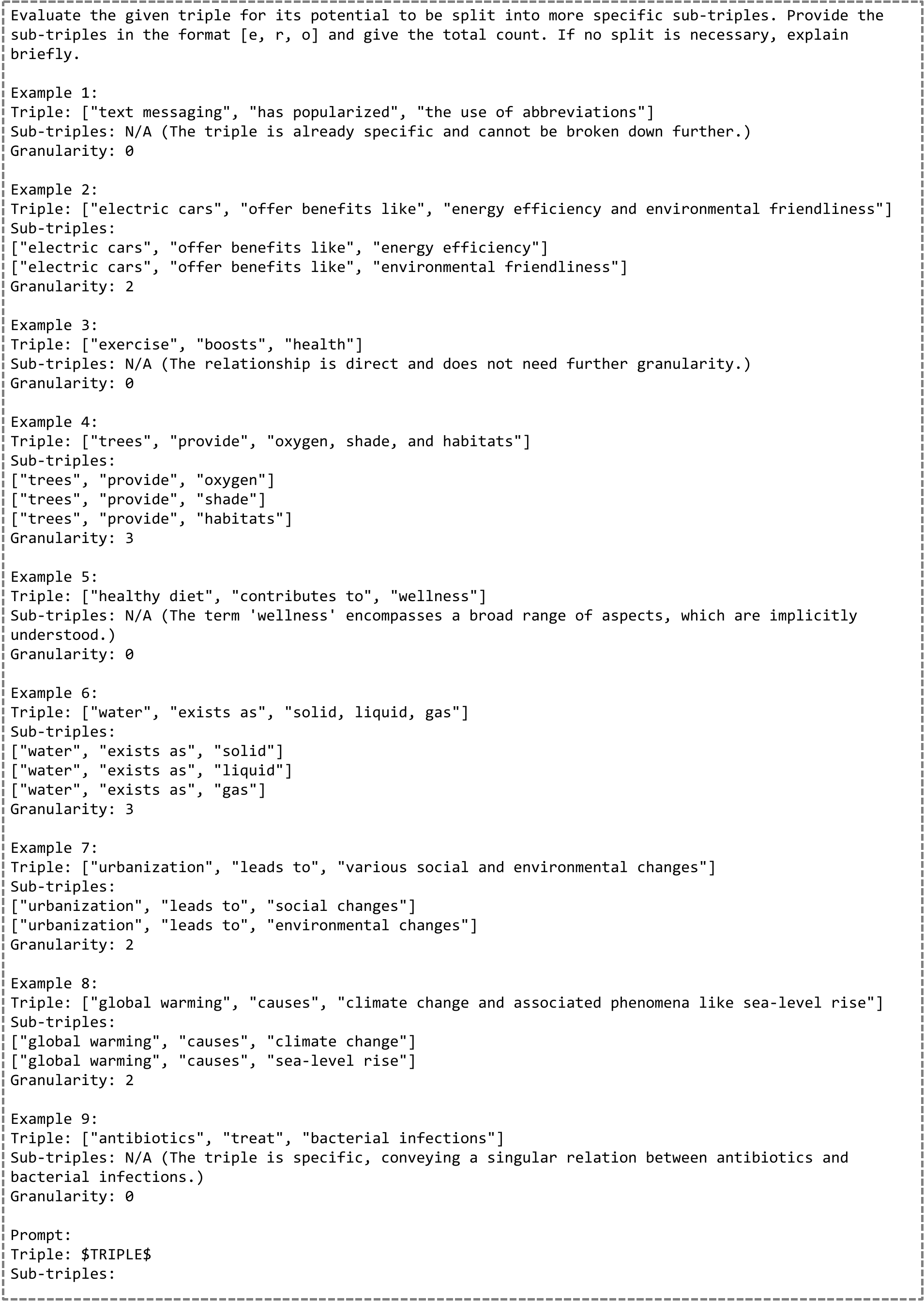}
    \caption{\textbf{Template for Granularity Checking.}}
    \label{fig:gran_prompt}
\end{figure*}




\end{document}